%% file: 0_main.tex
\documentclass[final]{llncs}
\usepackage{graphicx}
\usepackage{amsmath,amssymb} 
\usepackage{color}
\usepackage{tabularx}
\usepackage{makecell}
\usepackage{multirow} 
\usepackage{rotating}
\usepackage{floatrow}
\usepackage{wrapfig}

\def\isfinal{}

\ifx\isfinal\undefined
\usepackage{ruler}
\fi

\newcommand{\ie}{\textit{i.e. }}
\newcommand{\eg}{\textit{e.g. }}
\newcommand{\Name}{Syn2Real}

\newcommand{\veryshortarrow}[1][3pt]{\mathrel{%
   \hbox{\rule[\dimexpr\fontdimen22\textfont2-.2pt\relax]{#1}{.4pt}}%
   \mkern-4mu\hbox{\usefont{U}{lasy}{m}{n}\symbol{41}}}}

\usepackage[width=122mm,left=12mm,paperwidth=146mm,height=193mm,top=12mm,paperheight=217mm]{geometry}

\makeatletter
\renewcommand\subparagraph{%
  \@startsection{subparagraph}{5}
  {\parindent}
  {3.25ex \@plus 1ex \@minus .2ex}
  {-1em}
  {\normalfont\normalsize\bfseries}}
\makeatother
\usepackage{titlesec}
\let\subparagraph\relax
\titlespacing*{\section}{0pt}{0.7\baselineskip}{0.6\baselineskip}

\begin{document}
\pagestyle{headings}
\mainmatter
\def\ECCV18SubNumber{254}  

\title
{Syn2Real: A New Benchmark for Synthetic-to-Real Visual Domain Adaptation}
\titlerunning{ECCV-18 submission ID \ECCV18SubNumber}

\authorrunning{ECCV-18 submission ID \ECCV18SubNumber}

\ifx\isfinal\undefined

\author{Anonymous ECCV submission}
\institute{Paper ID \ECCV18SubNumber}

\else

\author{Xingchao Peng\textsuperscript{1$\star$}, Ben Usman\textsuperscript{1}\thanks{These authors contributed equally to this work.}, Kuniaki Saito\textsuperscript{2}, Neela Kaushik\textsuperscript{1}, \\Judy Hoffman\textsuperscript{3}, Kate Saenko\textsuperscript{1}}
\institute{Boston University\textsuperscript{1}, University of Tokyo\textsuperscript{2}, \\University of California Berkeley\textsuperscript{3}
}

\fi

\maketitle

\begin{abstract}
\vspace{-0.6cm}
  Unsupervised transfer of object recognition models from synthetic to real data is an important problem with many potential applications. The challenge is how to ``adapt'' a model trained on simulated images so that it performs well on real-world data without any additional supervision. Unfortunately, current benchmarks for this problem are limited in size and task diversity. In this paper, we present a new large-scale benchmark called \Name ~which consists of a synthetic domain rendered from 3D object models and two real-image domains containing the same object categories. We define three related  tasks on this benchmark: closed-set object classification, open-set object classification, and object detection.
  Our evaluation of multiple state-of-the-art methods reveals a large gap in adaptation performance between the easier closed-set classification task and the more difficult open-set and detection tasks. We conclude that
  developing adaptation methods that work well across all three tasks presents a significant future challenge for syn2real domain transfer. 
  
\end{abstract}

\input{1_introduction}
\input{2_related}
\input{3_visda_c}

\input{4_visda_d}
\input{5_openset.tex}
\input{6_conclusion.tex}

\clearpage

\bibliographystyle{splncs}
\bibliography{egbib}

\clearpage
\input{7_supplementary.tex}

\end{document}

%% file: 1_introduction.tex
\section{Introduction}

\begin{figure}[t]
    \centering
        \includegraphics[width=\linewidth]{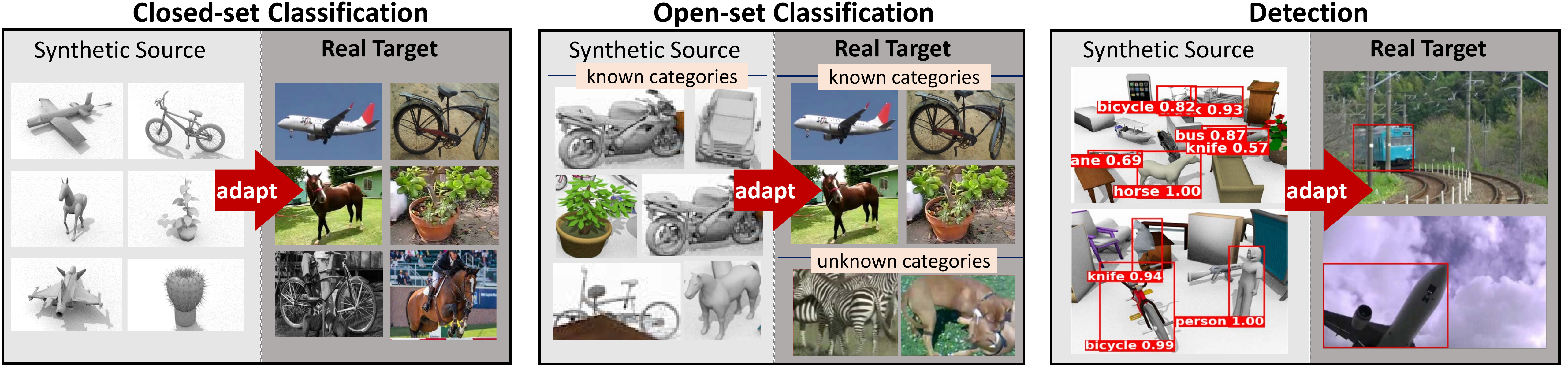}
    \vspace{-0.4cm}
    \caption{\scriptsize  (Best viewed in color) We present the \Name ~benchmark for unsupervised domain adaptation of object recognition models from a synthetic source domain to an unlabeled real target domain. The \Name~ Dataset includes multiple vision tasks all sharing the \textit{same synthetic-to-real domain shift}: traditional closed-set object classification, open-set classification with unknown object categories in the target domain, and object detection. 
    \vspace{-0.6cm}
    } 
    \label{fig:overview}
    \centering
\end{figure}

It is well known that the success of machine learning methods on visual recognition tasks is highly dependent on access to large labeled datasets. Unfortunately, model performance often drops significantly on data from a new deployment domain, a problem known as \textit{dataset shift} or \textit{dataset bias}~\cite{datashift_book2009}. Changes in the visual domain can include lighting, camera pose and background variation, as well as general changes in how the image data is collected. While this problem has been studied extensively in the domain adaptation literature~\cite{Csurka17}, progress has been  limited by the lack of large-scale challenge benchmarks. Existing cross-domain benchmarks ~\cite{nene1996columbia,office,carlucci2017,Tommasi14} created for evaluating domain adaptation algorithms often have limited size, low task diversity, and relatively small domain shifts.

The domain shift from simulated to real imagery can be particularly challenging. This shift occurs in many real-world situations where labeled imagery is difficult or expensive to collect but simulation is easy, such as in robotics, medical imaging, surveillance, etc. A synthetic rendering pipeline can produce virtually infinite amounts of labeled data, and can allow learning through interaction with the simulated environment 
\cite{Dosovitskiy17,richter2016playing}. CAD models of a wide variety of objects are freely available in online repositories. Yet existing benchmarks for synthetic-to-real adaptation either contain a very small number of object instances \cite{bousmalis2017unsupervised}, or focus on narrow tasks like eye tracking \cite{swirski2014rendering}.
Moreover, no existing benchmark defines multiple tasks on the same pair of synthetic and real domains.


In this paper, we introduce \textit{\Name}, a synthetic-to-real visual domain adaptation benchmark meant to encourage further development of robust domain transfer methods. The goal is to train a model on a synthetic ``source" domain and then update it so that its performance improves on a real ``target" domain, without using any target annotations. It includes three tasks, illustrated in Figure~\ref{fig:overview}: the more traditional \textit{closed-set classification} task with a known set of categories; the less studied \textit{open-set classification} task with unknown object categories in the target domain; and the \textit{object detection} task, which involves localizing instances of objects by predicting their bounding boxes and corresponding class labels. 


The \Name ~benchmark focuses on \textit{unsupervised} domain adaptation (UDA) for which the target domain images are not labeled. 
 While there may be scenarios where target labels are available (enabling \textit{supervised} domain adaptation), the purely unsupervised case is more challenging and often more realistic.
For each task, we provide a synthetic source domain and two real-image target domains.
Many UDA evaluation protocols do not have a validation domain and use labeled target domain data to select hyperparameters. However, assuming labeled target data goes against the UDA problem statement. For this reason, we collect two \textit{different} target domains, one for  validation of hyperparameters and one for testing the models.

For the closed-set classification task, we generate the largest synthetic-to-real dataset to date with over 280K images in the combined training, validation and test sets. We use this dataset to hold a public challenge, inviting researchers from all over the world to compete on this task, and analyze the performance of the top submitted methods. We find that for this task, where the object categories are known ahead of time, recent UDA models that leverage CNN features pre-trained on ImageNet are able to achieve impressive adaptation results. This is surprising, considering that labels are only available on synthetic source images and the source and target domains are very different. We provide a detailed analysis and insight into these results.

We then push the boundaries of synthetic-to-real transfer beyond closed-set classification and design an open-set classification dataset, where the target domains contain images of additional unknown categories that were not present in the source dataset. We evaluate the state-of-the-art UDA methods available for this more difficult task, and find that there is still much room for improvement. 

Furthermore, we propose an even more challenging object detection benchmark that covers a much more diverse set of object categories than previous syn2real detection datasets \cite{vazquez2014virtual,nexet}, and show that methods that excel for adaptation of classification models completely fail when applied to recent end-to-end detection models~\cite{liu2016ssd}, potentially due to very low initial performance of source models on the target data.

Our hope is that the \Name ~benchmark will provide a useful tool for the visual domain adaptation and transfer learning community. 
Besides enabling large-scale evaluation for the syn-to-real domain shift, it poses a challenge for the community to develop an adaptation method that works equally well for different tasks given the same domain shift.
We released our image datasets and benchmark code, and fully open sourced our 3D model collection, along with scripts and metadata required for proper rendering \footnote{http://ai.bu.edu/syn2real/}.

%% file: 2_related.tex
\section{Related Work}

There has been a lot of prior work on visual domain adaptation, ranging from the earlier shallow feature methods~\cite{office,BergamoTorresani10,duan10} to the more recent deep adaptation approaches~\cite{ganin2014unsupervised,Tzeng_2015_ICCV}.
A review of existing work in this area is beyond the scope of this paper; we refer the reader to a recent comprehensive survey~\cite{Csurka17}. 
Several most notable benchmark datasets that can be used to evaluate visual domain adaptation are summarized in Table~\ref{tab_datasets}. A majority of popular benchmarks lack task diversity: the most common cross-domain datasets focus on the image classification task, e.g., digits of different styles, objects~\cite{office} or faces~\cite{sim2002cmu} under varying conditions. 
Other tasks such as open-set classification~\cite{busto2017open}, detection~\cite{peng2015learning}, structure prediction \cite{RosCVPR16,Cordts2016Cityscapes,richter2016playing} and sequence labeling \cite{graves2012supervised} have been relatively overlooked.\\

\begin{table}[t]
    \centering
\scriptsize{
    \begin{tabular}{|c|c|ccccc|} \hline
    Dataset & Task & Examples & Classes & Domains & Syn$\veryshortarrow$Real & Multitask \\ \hline
    COIL20 \cite{nene1996columbia} & C & 1,440 & 20 (tools) & 1 & No & No \\
    Office \cite{office} &C & 1,410  & 31 (office) & 3 & No & No\\
    Caltech \cite{carlucci2017}& C & 1,123 & 10 (office) & 1 & No & No\\
    CAD-office \cite{peng2015learning} & C & 775 & 20 (office) & 1 & Yes & No \\

    
    Cross-Dataset \cite{Tommasi14} & C & 70,000+ & 40 (mixed) & 12 & No & No\\ 
    \hline
    CAD-PASCAL \cite{xiang_wacv14} & D+P & 12,000 & 20 (mixed) & 2 & Yes & Yes\\
    NEXAR \cite{nexet} & D & 55,000 & 1 (car) & 3 & No & No \\ 
    CVC \cite{vazquez2014virtual} & D & 5,616 & 1 (pedestrian) & 2 & Yes & No\\ \hline
    \textbf{\Name} & C+D+O  & 280,157 & 12 (mixed) & 3 & Yes & Yes \\ \hline
    \end{tabular} 
    }
    \caption{\scriptsize Comparison of \Name ~to existing cross-domain datasets used for domain adaptation experiments, with corresponding numbers of classes, samples, domains and tasks dataset was designed for: classification (C), detection (D), openset classification (O), pose esitmation (P). Although an openset classification dataset can be constructed from a classification dataset by merging a subset of categories into a single ``unknown'' category, as explored in \cite{busto2017open}, we claim that this setup artificially lowers diversity of the unknown set - the most challenging aspect of the openset setup. 
    \vspace{-0.6cm}
    } 
    \label{tab_datasets}
\end{table}

\noindent\textbf{Classification Datasets.} Adaptation of image classification methods has been among the most extensively studied problems of visual domain adaptation. One of the difficulties in re-using existing datasets to create multi-domain benchmarks is that the same set of categories must be shared among all domains. For digits (ten categories, 0-9), the most popular benchmark setup consists of three domains: MNIST (handwritten digits)~\cite{lecun1998gradient}, USPS (handwritten digits)~\cite{hull1994database} and SVHN (street view house numbers)~\cite{netzer2011reading}. Digit images are sometimes synthetically augmented to create additional domains, for example by inverting colors or using randomly chosen backgrounds \cite{ganin2014unsupervised}.
The Office dataset~\cite{office} is a popular benchmark for real-world objects. It contains 31 object categories captured in three domains: office environment images taken with a high quality camera (DSLR), the same environment captured with a low quality webcam (WEBCAM), and images downloaded from the amazon.com website (AMAZON). 

One problem with existing classification benchmarks is the relatively small domain shifts, such as the shift between two different sensors (DSLR vs Webcam in the Office dataset~\cite{office}), or between two handwritten digit datasets (MNIST vs USPS). Over time, improvements in underlying image representations and adaptation techniques have closed the domain gap on these benchmarks, and more challenging domain shifts are now needed to drive further progress.

Another issue with current datasets is their small scale. Modern computer vision methods require a lot of training data, while cross-domain datasets such as Office Dataset~\cite{office} only contain several hundred of images. The Cross-Dataset Testbed~\cite{Tommasi14} is a more recent classification benchmark. Its ``dense" version contains 40 classes extracted from Caltech256, Bing, SUN, and Imagenet with a minimum of 20 images per class in each dataset. It is significantly larger than Office, however, some domains are fairly close as they were collected in a similar way from web search engines. On the Caltech-Imagenet shift, adaptation performance has reached close to 90\% accuracy~\cite{carlucci2017}.

The popularity of the image classification task as a testbed may be due to lower effort required to engineer a good baseline model. Compared with other vision problems such as object detection, activity detection in video, or structured prediction, image classification is simpler and less computationally expensive to explore. Moreover, many state-of-the-art classification models are readily available for use as a baseline upon which adaptation can be applied. At the same time, other tasks have characteristics that present unique challenges for domain adaptation. In this work, we propose experimental setups for both the more common classification task, and the less studied open-set classification and detection tasks.

\noindent\textbf{Detection Datasets.} Compared to classification, cross-domain object detection is a far less studied problem. There are a few existing datasets that have been used to create multi-domain benchmarks for recognition and detection tasks in synthetic to real adaptation. Recent work in this area use adaptive learning from virtual scenarios to perform real-world pedestrian detection~\cite{vazquez2014virtual,vazquez2012unsupervised}. They use scenes from the CVC virtual-world pedestrian datasets to train models that predict on the Daimer AG real-world pedestrian detection dataset \cite{schneider2013pedestrian}. Beyond pedestrian detection, there has been some work done regarding cross-domain action detection in videos \cite{cao2010cross} and in facial detection (\eg facial action units) \cite{chu2013selective}. However, \Name ~appears to be among the first diverse object detection datasets designed specifically for adaptive learning.

\noindent\textbf{Open-set Datasets.} Open-set classification was considered in the transfer learning field only very recently. As such, there is a lack of open-set datasets designed specifically for domain adaptation tasks. One recent study \cite{busto2017open} generates benchmarks for open-set adaptation using the Office dataset and introduces an open-set protocol by adopting a set of 10 common classes from the Cross-Dataset Analysis \textit{Caltech} dataset to use as the known sets for classification. Beyond this introduced approach, however, there do not appear to be any other datasets for the open-set adaptation task. As in the detection task, \Name ~offers a valuable contribution to the visual domain adaptation community to study a novel problem.

\noindent\textbf{Synthetic-to-Real Datasets.} Synthetic data augmentation has been extensively employed in computer vision research. More specifically, 3D models have been utilized to generate synthetic images with variable object poses, textures, and backgrounds~\cite{peng2015learning}. Recent usage of 3D simulation has been extended to multiple vision tasks such as object detection~\cite{peng2015learning,sun2014virtual}, pose estimation~\cite{su2015render}, robotic simulation~\cite{tzeng2015towards}, semantic segmentation~\cite{richter2016playing}. Popular 3D model databases of common objects that may be used in visual domain adaptation tasks include  ObjectNet3D \cite{xiang2016objectnet3d}, ShapeNet and the related ModelNet \cite{shapenet2015}. Table~\ref{tab_synth_datasets} in the supplementary appendix compares the \Name ~dataset to existing synthetic object datasets.

%% file: 3_visda_c.tex
\section{\Name ~Benchmark}

We introduce the \textit{\Name} dataset as a large-scale testbed for studying unsupervised domain transfer and adaptation of object category models. We collect 3D models and real images of the same 12 object categories, resulting in three visual domains:
\vspace{-2mm}
\begin{itemize}
\item \textbf{training domain (source):} synthetic renderings of 3D models from different angles and with different lighting conditions,
\item \textbf{validation domain (target):} a real-image domain consisting of images cropped from the Microsoft COCO dataset~\cite{mscoco14},
\item \textbf{testing domain (target):} a real-image domain consisting of images cropped from the Youtube Bounding Box dataset~\cite{real17youtube}.

\end{itemize}

The goal is to train a vision system on images from a synthetic domain and then improve its performance on a real target domain. We propose to utilize different target domains for the validation and test splits to prevent hyper-parameter tuning on the test data. Unsupervised domain adaptation is usually done in a \textit{transductive} manner, meaning that unlabeled test data is actively used to train the model. However, it is not possible to tune hyper-parameter on the test data, since it has no labels. Despite this fact, the lack of established validation sets often  leads to  experimental protocols where the labeled test set is used for this purpose. In our benchmark, we provide a validation set to mimic the more realistic deployment scenario where the target domain is unknown at training time and test labels are not available for hyper-parameter tuning. This setup also discourages algorithms that are designed to handle a specific target domain. The validation and test sets are \textit{different} domains, so over-tuning to one can potentially degrade performance on another.

The \Name ~benchmark consists of three tasks, each defined across the above three domains: closed-set classification (\Name-C), open-set classification (\Name-O), and detection (\Name-D). The details of the data collection and experimental results of each of these tasks are described below.

\begin{table}[t]

\scriptsize
{

\noindent
\begin{tabular}{p{1.8cm}| p{0.6cm} p{0.5cm} p{0.65cm} p{0.65cm} p{0.6cm} p{0.6cm} p{0.75cm} p{0.8cm} p{0.6cm} p{0.65cm} p{0.6cm} p{0.6cm}  p{0.65cm} | p{0.65cm}     }
\Xhline{0.6pt}
   Number & aero  & bike &bus &car & horse &knife & mbike & person& plant& skbrd & train & truck &other    & Total\\
   
\hline
   of models & 179 & 93 & 208& 160& 119& 178 & 217 & 152 & 135 & 146 & 200 & 120 & - &1,907 \\
\hline
 Training (C) & 14.3k & 7.4k & 16.6k & 12.8k & 9.5k & 14.2k & 17.4k & 12.1k & 10.4k & 11.7k & 16k & 9.6k& -& 152k \\
 Validation (C) & 3.6k& 3.5k& 4.7k & 10.4k & 4.7k & 2.1k & 5.8k & 4.0k& 4.5k & 2.3k & 4.3k &5.5k & - & 55.4k\\
 Testing (C) &5.2k & 4.3k & 7.0k & 7.3k & 6.3k & 5.5k & 8.1k & 7.7k & 4.3k & 2.8k & 7.3k & 6.8k & - &72.3k \\

\hline
Training (O) & 6.0k & 6.6k & 6.1k& 6.1k& 7.2k& 6.6k& 6.5k& 7.5k&7.1k& 6.0k& 6.2k&6.4k& 64.5k &  143k\\ 
Validation (O) & 3.6k & 3.5k& 4.6k& 10.4k& 4.7k& 2.1k& 5.8k&4.0k& 4.5k& 2.3k&4.2k& 5.5k& 49.7k & 105k\\

\Xhline{0.6pt}

\end{tabular}

} 

\caption{\scriptsize Number of images in the domains of the \Name ~closed-set classification (C) and open-set classification (O) tasks. The \Name-O dataset is generated from the detection (D) dataset, which consists of 10k images with objects from 12 known categories and 33 unknown categories
\vspace{-0.6cm}
}
\label{tab_num_of_models_and_images}
\end{table}



\section{\Name-C: Closed-set Classification Task}
\begin{figure}[p]
    \centering
    \includegraphics[width=\linewidth]{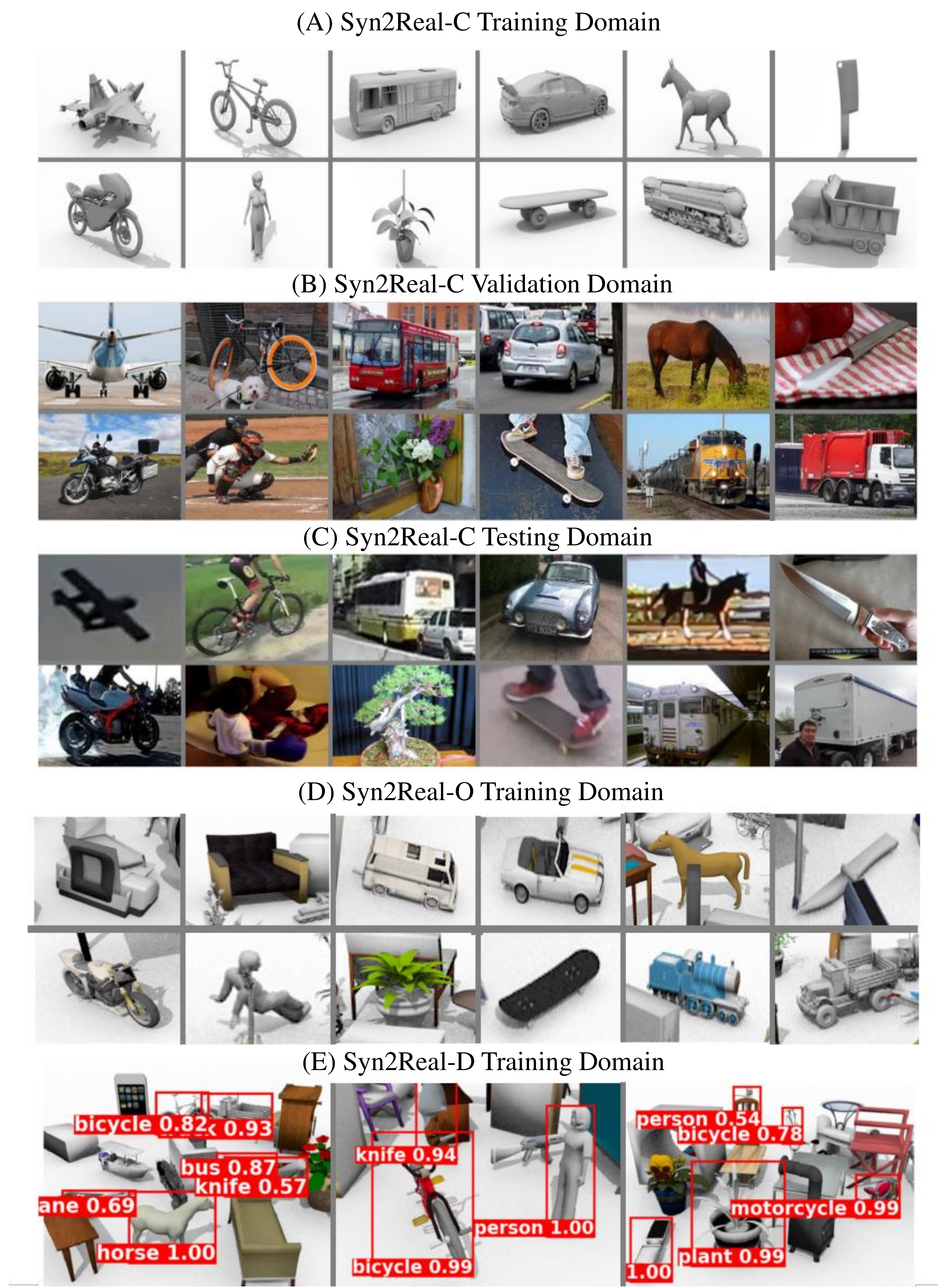}
    \caption{\scriptsize Sample images from the Syn2Real dataset. Group (A) shows synthetically rendered \Name-C images (source domain); group (B) and group (C) show \Name-C images cropped from COCO dataset~\cite{mscoco14} using their bounding boxes (validation target domain) and similarly cropped images from YouTube-BB dataset~\cite{real17youtube} (test target domain), respectively. Two groups on the bottom show samples from source openset dataset \Name-O including some ``unknown'' objects such as a printer and a chair, and source detection dataset \Name-D with ground truth bounding boxes and occlusion rates. Please refer to supplemental material for additional visualizations.}
    \label{fig_cls_sample}
    \centering
\end{figure}

Our first task is to learn to classify images from the synthetic domain, and transfer the learned models to a target real domain. For this task, we assume that the target domain images belong to the same set of known categories as the source.
We collect the largest-scale dataset to date for this problem; the number of images in each category in \Name-C is shown in Table~\ref{tab_num_of_models_and_images} and Figure \ref{fig_cls_sample} shows some examples. 

\vspace{-0.5cm}

\subsection{Dataset Acquisition}
\vspace{-0.2cm}

\noindent \textbf{Training Domain: CAD-Synthetic Images}. We generate the source domain by rendering 3D models of the same 12 object categories as in the real data from different angles and under different lighting conditions. We obtained 1,907 3D models in total from several free online sources, the details of which are described in the supplemental material. 
Twenty different camera yaw and pitch combinations with four different light directions are used to render images from each model. The lighting setup consists of ambient and sun light sources in 1:3 proportion. Objects were rotated, scaled and translated to match the floor plane, duplicate faces and vertices were removed, and the camera was automatically positioned to capture the entire object with a margin around it. For textured models, we also rendered their un-textured versions with a plain grey albedo. In total, we generated 152,397 images to form the synthetic source domain.

\noindent \textbf{Validation Domain: MS COCO.} The validation dataset for \Name ~is built from the Microsoft COCO dataset \cite{mscoco14}. 
In total, the MS COCO dataset contains 174,011 images. We used annotations provided by the COCO dataset to find and crop relevant object in each image. All images were padded by retaining an additional 50\% of its cropped height and width. Padded image patches whose height or width was under 70 pixels were excluded from the dataset.
In total, we collected 55,388 object images that fall into the chosen twelve categories.  
We took all images from each of twelve categories with the exception of the ``person" category, which was limited to 4,000 images in order to balance the overall number of samples per category (the original ``person''  category had more than 120k images). 

\noindent \textbf{Testing Domain: YouTube Bounding Boxes.} Due to the high number of object category labels overlapping with the other two domains, we chose the YouTube Bounding Boxes (YT-BB) dataset~\cite{real17youtube} to construct the test domain. Compared to the validation domain (MS COCO), the resolution of images in YT-BB is much lower, because they are frames extracted from YouTube videos. The original YT-BB dataset contains segments extracted from 240,000 videos and approximately 5.6 million bounding box annotations for 23 categories of tracked objects. We extracted 72,372 frame crops that fall into one of our twelve categories and satisfy the size constraints. 

\subsection{Experiments}
\label{sec_experimental_setup}
Our first set of experiments aims to provide a set of baselines for the \Name-C benchmark. Below we also present and analyze the results obtained by the winners of the challenge we held on this dataset.

\noindent\textbf{Experimental Setup.}  We perform in-domain (\ie train and test on the same domain) experiments to obtain  approximate ``oracle'' performance, as well as source-only (\ie train only on the source domain) to obtain the lower bound results of no adaptation.
In total, we have 152,397 images as the source domain and 55,388 images as the target domain for validation. In our in-domain experiments, we follow a 70\%/30\% split for training and testing, i.e., 106,679 training images, 45,718 test images for the synthetic domain and 38,772 training images, 16,616 test images for the real domain. 

We first adopt the widely used AlexNet CNN architecture as the base model. The last layer is replaced with a fully connected layer with output size 12. We initialize the network with parameters learned on ImageNet~\cite{ILSVRC15}, except the for the last layer, which is initialized with random weights from $\mathcal{N}(0, 0.01)$. We utilize mini-batch stochastic gradient descent (SGD) and set the base learning rate to be $10^{-3}$, weight decay to be $5 \times 10^{-4}$ and momentum to be 0.9. We also utilize ResNext-152~\cite{resnext} as a base model; the output dimension of the last fully connected layer is changed to 12 and initialized with ``Xaiver'' parameter initializer~\cite{glorot2010understanding}. Since the output layer is trained from scratch, we set the learning rate to be 10 times that of other layers. The learning rate is adjusted with the formula: $\eta_{p} = \frac{\eta_{0}}{(1+\alpha p)^{\beta}}$, where $p$ will linearly change from 0 to 1 along the training process, $\eta_{0} = 10^{-4}$, $\alpha = 10$ and $\beta = 0.75$. The latter setup gave highest source-only results. We report the  accuracy of classification at 40k iterations.

\noindent\textbf{Domain Adaptation Baselines.} We evaluate two existing domain adaptation algorithms as baselines. \textbf{\textit{DAN}} (Deep Adaptation Network)~\cite{long2015learning} learns transferable features by training deep models with Maximum Mean Discrepancy~\cite{sejdinovic2013equivalence} loss to align the feature distribution of source domain to target domain. In our implementation, the network architecture of \textbf{\textit{DAN}} is extended from AlexNet~\cite{alexnet}, which consists of 5 convolutional layers (\textit{conv1 - conv5}) and 3 fully connected layers (\textit{fc6 - fc8}) and \textbf{\textit{Deep CORAL}} (Deep Correlation Alignment)~\cite{sun2015return}  performs deep model adaptation by matching the second-order statistics of feature distributions. The domain discrepancy is then defined as the squared Frobenius norm $d(S, T) = \lVert \operatorname{Cov}_{S}-\operatorname{Cov}_{T}\rVert_{F}^2$, where $\operatorname{Cov}_{S}, \operatorname{Cov}_{T}$ are the covariance matrices of feature vectors from the source and target domain, respectively. 


\begin{table}[t]
\centering
\vspace{-0.3cm}
\scriptsize
{

\textit{\textbf{Training Domain (CAD-synthetic) $\rightarrow$ Validation Domain (MS COCO)}}
\noindent
\begin{tabular}{p{1.4cm}| p{0.4cm}p{0.5cm}|p{0.45cm} p{0.45cm} p{0.45cm} p{0.45cm} p{0.45cm} p{0.45cm} p{0.45cm} p{0.45cm} p{0.45cm} p{0.45cm} p{0.45cm} p{0.45cm}  !{\vrule width0.8pt} p{0.6cm} |p{0.8cm} | p{0.9cm}   }
\Xhline{0.8pt}
 Method   &\rotatebox{60}{Train} &\rotatebox{60}{Test} & \rotatebox{60}{aero}  & \rotatebox{60}{bike} & \rotatebox{60}{bus} & \rotatebox{60}{car} & \rotatebox{60}{horse} & \rotatebox{60}{knife} & \rotatebox{60}{mbike} & \rotatebox{60}{person} & \rotatebox{60}{plant} & \rotatebox{60}{skbrd} & \rotatebox{60}{train} & \rotatebox{60}{truck}  & \rotatebox{60}{Mean} & \rotatebox{60}{Source} & \rotatebox{60}{Gain}\\
 \Xhline{0.8pt} 

DAN   & syn & real & 71 &	47 & 67 & 31 & 61 &	49 & 72 & 36 & 64 &	28 & 70 & 19 & 51.6 & 28.1 &  83.6\%\\
D-CORAL & syn & real & 76 & 31 & 60 & 35 &	45 & 48 & 55 & 28 & 56 & 28 & 60 & 19 & 45.5 & 28.1& 61.9\% \\
BUPT & syn & real & 96&	70&	83&	56&	95&	88&	95&	71&	88&	69&	95&	28&	77.8 & 55.8 & 39.4\% \\
SE & syn & real & 97 & 87 & 84 & 64 &	95 & 96 & 92 & 82 & 96 & 92 & 87 & 54 & 85.5& 43.0 & 98.8\% \\
SE* & syn & real & 4 & 0 & 3 & 0 & 3 & 70 & 3& 6 & 29 & 1 & 0 & 5 & 10.4 & - &-\\
\hline
Source & syn & real & 53 & 3 & 50 & 52 & 27 & 14 & 27 & 3 & 26 & 10 & 64 & 4 & 28.1 & - & - \\
Oracle & syn & syn & 100 & 100 & 99 & 99 & 100 & 99 & 99 & 100 & 100 & 100	& 99 & 99 & 99.9 &-&- \\
Oracle &real& real& 94& 83 & 83& 86& 93 & 91 &	90 & 86 & 94 &	88 & 87 &	65 & 87.2&-&- \\
\Xhline{0.8pt}
\end{tabular}

\vspace{0.1cm}
\textit{\textbf{Training Domain (CAD-synthetic) $\rightarrow$ Testing Domain (YT-BB)}}
\begin{tabular}{p{1.4cm}| p{0.4cm}p{0.5cm}|p{0.45cm} p{0.45cm} p{0.45cm} p{0.45cm} p{0.45cm} p{0.45cm} p{0.45cm} p{0.45cm} p{0.45cm} p{0.45cm} p{0.45cm} p{0.45cm}  !{\vrule width0.8pt} p{0.6cm} |p{0.8cm} | p{0.9cm}   }
 
\Xhline{0.8pt}
 
DAN&syn&real &55 & 18 & 60 & 69 &55 & 41 & 63 & 30 & 79 & 23 & 63 & 40 & 49.8 & 30.8 & 61.6\%\\
D-CORAL& syn&real& 63 & 22 & 66 & 65 & 31 & 37 & 54 & 25 &	74 & 30 & 43 & 34 & 45.3 & 30.8 & 47.0\%\\
BUPT & syn& real& 96& 67& 93& 97& 91& 87& 92& 74& 96& 67& 95& 69& 85.4& 63.2& 35.1\%\\
NLE\_DA &syn&real& 94& 86& 87& 95& 91& 90& 82& 78& 96& 77& 87& 88& 87.7& 64.3& 36.4\%\\
SE&syn&real& 97 & 91 & 97 & 95 &	98 & 82 & 73 & 98 & 93 & 94 & 60 & 94 & 91.8 & 45.5 & 102\% \\
SE* & syn & real & 6&2 &3& 5& 1& 66 & 54 & 5 &4 &0 & 1 & 4 & 12.6& - & - \\

\hline
Source &syn& real& 46 &1 & 59 & 83 & 21 & 14 & 23 & 1 & 46 & 17 & 47 & 10 & 30.8 & - & - \\
Oracle & real & real & 95 & 84 & 90 & 96 & 93 & 95 & 90 & 90 & 96 & 89 & 95 & 92 &  92.1 &  - & - \\
Oracle-Res & real & real & 96 & 89 & 93 & 98 & 95 & 96 & 91 & 92 & 96 & 86 & 95 & 94 & 93.4 & - & - \\

\Xhline{0.8pt}
\end{tabular}
} 

\caption{\scriptsize \textbf{Baseline and challenge results for the closed-set classification track (\Name-C)} on the \textit{validation} domain (top table) and \textit{test} domain (bottom table). Columns show:  method name, train/test data, per-category accuracy, mean per-category accuracy (Mean), mean per-category accuracy without adaptation (Source) and the relative gain obtained by adaptation (Gain). SE* uses no CNN pretraining on ImageNet. See text for more details. 
\vspace{-0.6cm}
}\label{tab_cls}
\vspace{-0.2cm}
\end{table}

\noindent\textbf{Baseline Results.} Baseline results on the validation domain for classification are shown in Table~\ref{tab_cls}. ``Oracle" or in-domain AlexNet performance for training and testing on the synthetic domain reaches 99.9\% accuracy, and oracle training and testing on the real validation domain leads to 87.6\%. This supervised learning performance provides a loose upper bound for our adaptation algorithms. As far as unadapted source-only results on the validation dataset, AlexNet trained on the synthetic source domain and tested on the real domain obtains 28.1\% accuracy, a significant drop from in-domain performance. This provides a measure of how much the domain shift affects the model. Among the tested domain adaptation algorithms, Deep CORAL improves the cross-domain performance from 28.1\% to 45.5\% and DAN further boosts the result to 51.6\%. While their overall performance is not at the level of in-domain training, they achieve large relative improvements over the base model through unsupervised domain adaptation, improving it by 83.6\% and 61.9\% respectively.

On the test domain (Table~\ref{tab_cls} bottom half), AlexNet achieved similar oracle and source-only performance to that on the validation dataset.  Oracle performance of AlexNet is 92.1\% and ResNext-152 improves the result slightly to 93.4\%. Source AlexNet achieves 30.8\% mean accuracy, and DAN and Deep CORAL improve the result to 49.8\% and 45.3\%, respectively. As a base model, AlexNet has relatively low performance due to its simpler architecture, compared to more recent CNNs. However, the relative improvement of domain adaptation algorithms (\ie DAN and Deep CORAL) is still large. The source model based on ResNet-152 gives higher source accuracies then Alexnet-based and was used by all winning teams.


\noindent\textbf{Challenge results.} We posted the \Name-C dataset and evaluation kits online and invited the public to participate in a challenge. Table~\ref{tab_cls} shows the top three of the many submitted domain adaptation results. The top performing 
team used the Self-Ensembling (SE in the table) algorithm \cite{french17self}. It improved their source-only ResNet-152 model from 45.3\% to 91.8\%, a 102\% relative improvement. Authors use two networks with same architecture: student and teacher networks, and the latter has its weights set to an \textit{exponential moving average of weights} of the student network on previous training iterations. The method consisted of optimizing three losses: 1) a cross entropy between ground truth and predictions of the so-called student network on samples from the source domain, 2) a mean square difference between predictions of student and teacher networks on samples from target domain for which prediction confidence exceeds a threshold, and 3) cross entropy between uniform distribution and distribution of predicted target class labels. Therefore optimizing an agreement of the student network trained in a supervised fashion with its copies from the past on samples from target for which prediction confidence is high results in essentially bootstrapping target labels with high confidence predictions. Moreover, the resulting loss is trained in minibatches with dropout, noise and random data augmentations to improve robustness of the resulting procedure. 

The second and third-best performing teams, \textit{NLE\_DA} and \textit{BUPT\_OVERFIT}, made use of maximum mean discrepancy (MMD) with different bandwidth priors measured between single resulting feature representation or combined representations from multiple layers (JMMD). Late fusion of features extracted from off the shelf deep models pretrained on Imagenet resulted in a substantial improvement in terms of the test score. 

\vspace{-0.5cm}
\subsection{Analysis of Results}
\vspace{-0.2cm}

The challenge participants obtained excellent performance in our \Name-C test domain, on par with in-domain training. The self-ensembling model (SE) achieved particularly good performance. In this section, we provide a more in-depth analysis of its results. We investigate how the following factors affect its performance: number of synthetic training images, inclusion of small/irregular images in the target data, 
adding synthetic background clutter to synthetic images,
and pre-training the CNN on ImageNet.

\begin{wrapfigure}{r}{0.5\textwidth}
\vspace{-0.5cm}
    \centering
    \includegraphics[width=\linewidth]{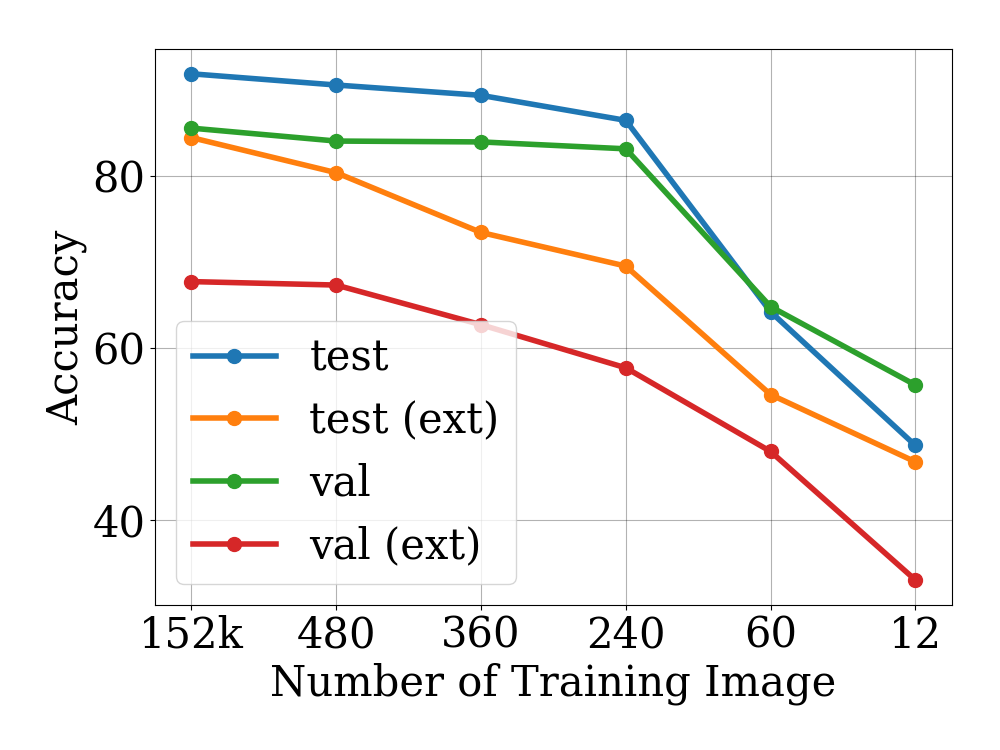}
    \caption{\small We examined how accuracy of the adapted SE classifier is affected by reducing the number of training images in the source domain (from 152K to 12), and inclusion of small/irregular images to the test set (ext).     \vspace{-1cm}}
    \label{fig_accuracy_ext}
    \centering
\end{wrapfigure}

\noindent \textbf{Effect of number of synthetic images.} We performed an ablation study to understand the effective number of synthetic images required for successful adaptation. Figure~\ref{fig_accuracy_ext} shows the mean per-class accuracy of the SE model for the original 152K training set and for significantly smaller sets (\textit{test} and \textit{val} curves). Accuracy remains stable when using only 480 source images, and starts to drop significantly at 240 images. We conclude that on the \Name-C target domains, SE with Imagenet pre-training does not require a lot of synthetic data.

\noindent \textbf{Effect of target image size.} We generate an extended version of the target domains referred as \Name-C-ext by also including target images whose height or width are under 70 pixels, to increase the difficulty of the task. In total, we add 35,591 images to the MS-COCO val domain and 4,533 images to the YT-BB test domain. Accuracy on \textit{val(ext)} and \textit{test(ext)} in Figure~\ref{fig_accuracy_ext} drops significantly compared to the original domains, especially on \textit{val(ext)} (MS-COCO). We also observe that accuracy drops off faster with smaller training set size on these more difficult target domains.

\noindent \textbf{Effect of synthetic background.} We also ran classification experiments on images corresponding to twelve known classes cropped from the \Name-D detection dataset, as they include some background clutter, to see if that would make source images more similar to the target and therefore reduce domain shift. We observed a 3-5\% drop in target performance on majority of categories and a 10-20\%  drop on ``knife'', ``skateboard'' and ``truck'' categories. We suggest that this must be due to our synthetic background not conveying any additional information about object category, whereas in real images the background is often correlated with object class (e.g., sheep appear on green grass, etc.)

\noindent \textbf{Effect of pre-training.}
All of the adaptation models so far have initialized the source CNN models with features pre-trained on the ImageNet dataset. As we saw above, with such strong starting features, the SE adaptation model can achieve good performance even with very few synthetic images. We also tried using random initialization (no pre-training) in several domain adaptation models but the results were below 20\% accuracy.
The adapted performance of the SE model not initialized with Imagenet features is reported as SE* in Table~\ref{tab_cls}. Its performance highly depends on the number of synthetic images and does not saturate, however overall accuracy is very low.

%% file: 4_visda_d.tex
\section{\Name-D: Detection Task}

We define an object detection task on the \Name. The goal is to localize each object from one of the 12 categories and predict its class and its bounding box. 

\noindent\textbf{Dataset Acquisition.}
We use 3D CAD models collected and annotated for \Name-C as well as models from 33 additional categories from ShapenetCore to generate ten thousand images for the detection task, each having around ten known and ten unknown objects. Here the ``known'' categories are those we want to detect, whereas the ``unknown'' are those that appear in the background. The rendering pipeline is similar to the one described in the previous section, apart from models being scaled and then located semi-randomly across the scene. Object bounding boxes were inferred from instance segmentation maps obtained by replacing material fragment shader with a constant value shader. Occlusion rates were obtained similarly by comparing a whole scene instance segmentation map to an instance segmentation maps of the scene with all objects but one removed, an illustrating example is given in the supplementary. We used first 10K images from COCO and YTBB detection datasets as the target. 

\noindent \textbf{Baseline experiments.} 
As to our knowledge there are no published baselines considering adaptation for detection of multiple object types using end-to-end detection pipelines such as SSD \cite{liu2016ssd}, Faster-RCNN \cite{renNIPS15fasterrcnn} or YOLO \cite{redmon2016yolo9000}, therefore we propose the following basic setup. We trained an SSD model consisting of a VGG16 feature extractor pretrained on Imagenet and a region proposal network predicting class labels, locations and size shifts for a total of 8k possible anchor boxes using feature maps from five last convolutional layers. We chose SSD (not YOLO or Faster-RCNN) because of its relative simplicity: if SSD does not perform well, then more compilcated models that use same building blocks will also fail. We applied multiple domain adaptation methods originally proposed for classification models to the resulting detection model as shown in Table \ref{tab_detection}. Columns of the table refer to the mean average precision of the bounding boxes that have intersection over union with ground truth bounding boxes higher then 0.5, localization and classification components of the loss, as well as mean distance between feature representations computed by corresponding adaptation methods. We computed Maximum Mean Discrepancy (MMD) between outputs of the model on source and on target as well as between the last three convolutional feature representations, whereas CORAL was computed on last convolutional feature representations considering activations for each spatial region as an individual observation when computing covariance matrices. We also used the self-ensembling method for detection by
minimizing difference between output of the student model that is trained with gradient descent and a teacher model with weights equal to the exponential average of student's weights. This difference was only considered for predictions with label confidence exceeding 0.9 threshold. We also added a class balance loss that encouraged same distribution of output classes for source and target input batches. Table \ref{tab_detection} shows results of these experiments as well as performance of the model trained on source and right after Imagenet initialization. The student-teacher loss is equal to zero on the first iteration of SE because no predictions exceed the confidence threshold. By the end of training quite a few prediction do, and the class balance loss is also much lower.

\noindent \textbf{Discussion.} Even though the discrepancy measures between feature representations such as CORAL and SE losses decreased over time, target performance did not improve. We suggest that this must be due to very low (close to zero) performance of the source model on target, and therefore bootstrapping is more challenging from a poor model. This agrees with our no-pretraining results for the classification task. Moreover, feature representations given by large convolutional maps are very high-dimensional, so conventional measures such as MMD and CORAL might not be a good fit here, as these methods are usually applied to outputs of the last fully connected layers which have only a few thousand dimensions. We conclude that the detection task presents a significant future challenge for domain adaptation methods. The covariance matrices for CORAL loss were computed considering activations for each spatial region of the output as an individual observation.

\begin{table}[t]
\centering
\scriptsize
{
\vspace{-0.3cm}
\begin{tabular}{l| c c c | c c c | c c}
\Xhline{0.8pt}
\multirow{2}{*}{Method} & \multicolumn{3}{c|}{Source} & \multicolumn{3}{c|}{Target} & \multicolumn{2}{c}{Discrepancy} \\
& mAP & $\mathcal L_{\text{loc}}$ & $\mathcal L_{\text{class}}$ & mAP & $\mathcal L_{\text{loc}}$ & $\mathcal L_{\text{class}}$ & $\mathcal L_D^{(0)}$ & $\mathcal L_D^{(T)}$\\
 \Xhline{0.8pt} 
Initialization & $10^{-3}$ & 3.0 & 13.5 & $10^{-4}$ & 3.5 & 14.0 & \multicolumn{2}{c}{\multirow{3}{*}{n\textbackslash a}} \\
Source only & 0.76 & 0.3 & 1.3 & 0.06 & 3.1 & 6.8 & &  \\
Target only & 0.26 & 1.4 & 3.8 & 0.35 & 1.2 & 2.2\\
\hline
MMD & 0.60 & 0.9 & 2.2 & 0.03 & 4.1 & 8.0 & 1$\cdot10^{-7}$ & 1$\cdot10^{-5}$ \\
CORAL & 0.42 & 1.5 & 3.3 & 0.04 & 3.5 & 5.0 & 5$\cdot10^4$ & 0.4 \\
SE& 0.58 & 0.9 & 2.4 & 0.02 & 3.2 & 5.5 &  (0, 0.27) & (0.04, 0.06)\\
\Xhline{0.7pt}
\end{tabular}

} 

 \vspace{-0.2cm}
\caption{\scriptsize Experiment results for the detection adaptation task using SSD300 with VGG16 feature extractor. We report mean average precision (mAP) at 0.5 IOU on source (synthetic renders) and target (COCO17-val), localisation ($\mathcal L_{\text{loc}}$) and classification ($\mathcal L_{\text{class}}$) losses, and domain discrepancy loss at the beginning $\mathcal L_D^{(0)}$ and at the end $\mathcal L_D^{(T)}$ of adaptation procedure. For SE method we report a pair of masked student-teacher loss and class balance loss.
\vspace{-0.8cm}}
\label{tab_detection}
\end{table}

%% file: 5_openset.tex
\section{\Name-O: Open-set Classification Task}

Open-set domain adaptation considers  classification  when the target domain contains categories unknown (unlabelled) in the source domain. It can be viewed as an intermediate task between closed-set classification and detection.

\noindent\textbf{Dataset Acquisition.} We cropped object instances from the \Name-D dataset described in the previous section using bounding boxes of all objects including the 33 ``background'' objects to build an open-set classification dataset. The 12 original categories are considered ``known'' and the 33 background categories are ``unknown''. We only chose objects that were 
more than 60\% non-occluded. 
The target domain was constructed from COCO using known images from the validation split of \Name-C and 50k images from 69 other COCO categories (see Table \ref{tab_num_of_models_and_images} for details).

\noindent\textbf{Baseline experiments.}
In a recent study on open-set adaptation, \cite{busto2017open} suggested a shallow semi-supervised approach based on iteratively solving an assignment problem and aligning two domain instances on subsets of the Office dataset. We explored what we can obtain with  more conventional deep distribution matching methods. These methods can be applied to open set domain adaptation when unknown source samples are available. The underlying idea is to match the distribution of unknown source samples with that of unknown target ones, even though the categories of unknown source and target are different. MMD~\cite{long2015learning} and BP~\cite{ganin2014unsupervised} are utilized as baseline methods in~\cite{busto2017open}. In addition, several methods for open set recognition have been proposed for the case where unknown samples are not provided during training~\cite{jain2014multi,bendale2015towards}. Open-set SVM~\cite{jain2014multi} rejects unknown samples lower than a pre-defined probability threshold, and works well when known and unknown samples belong to the same domain. Recent AODA~\cite{aoda2018saito} method utilizes adversarial approach for both domain adaptation and unknown outlier detection.

We prepared six baselines, including Source Only and distribution matching based methods. In addition, we applied self-ensembling based method (SE)~\cite{french17self} to this setting. 
With regard to the six baselines, we trained models with existing distribution alignment methods as was done in the closed-set task. We present the results of Source Only, MMD and DANN based on ResNet152 that was pre-trained on ImageNet. 
The top fully-connected layer of the network was removed and replaced with three fully-connected layers with batch normalization~\cite{ioffe2015batch} and a Leaky ReLU activation layer. We update the parameters of these newly added layers and fix the pre-trained parts of the network. In addition, we apply Open-set SVM~\cite{jain2014multi} to the features obtained by the methods. Open-set SVM does not need unknown samples for training, so we train it only on known source samples and test it on target samples, rejecting unknown samples lower than 0.1 in all experiments. 
To obtain the result of Oracle, we trained the networks in the same way as above. 

\input{table_5.tex}

\noindent\textbf{Ablation Study.}
To investigate how the known-to-unknown ratio affects experimental results, we change the known-to-unknown ratio of target domain to 1:10, but keep the ratio of source domain. The experimental results listed in Table~\ref{tab_cls_opn} show that the accuracy drops significantly after the known-to-unknown ratio is set to 1:10.

\noindent \textbf{Discussion.}
Table \ref{tab_cls_opn} shows the results of the experiment. We can see the effectiveness of adaptation by comparing Source Only with MMD, DANN and SE. By applying Open-set SVM to each method, a slight improvement was observed, even though Open-set SVM does not utilize unknown samples. It explains why it has lower unknown accuracy. The accuracy was very low for some classes such as bike, person and truck. Unknown target samples seem to prevent from correctly aligning known target samples with known source samples. The mean per-category accuracy of SE is higher than other baselines, however, the overall accuracy was actually worse than others, because the SE method did not perform well at recognizing unknown objects on the target, as the table demonstrates. This also agrees with our detection experiment results, where the SE method seemed to have degraded the source model performance by not adequately handling negative boxes. 


%% file: table_5.tex
\begin{table}[t]
\centering
\vspace{-0.3cm}
\scriptsize
{

\textit{\textbf{Training Domain (CAD-synthetic) $\rightarrow$ Validation Domain}}\\
\textit{\textbf{Known-to-Unknown Ratio = 1:1}}
\noindent

\begin{tabular}{p{1cm}| p{0.4cm}p{0.5cm}|p{0.45cm} p{0.45cm} p{0.45cm} p{0.45cm} p{0.45cm} p{0.45cm} p{0.45cm} p{0.45cm} p{0.45cm} p{0.45cm} p{0.45cm} p{0.45cm} p{0.45cm} !{\vrule width0.8pt} p{0.6cm} p{0.6cm} p{0.6cm} }
\Xhline{0.8pt}
 Method   &\rotatebox{60}{Train} &\rotatebox{60}{Test} & \rotatebox{60}{aero}  & \rotatebox{60}{bike} & \rotatebox{60}{bus} & \rotatebox{60}{car} & \rotatebox{60}{horse} & \rotatebox{60}{knife} & \rotatebox{60}{mbike} & \rotatebox{60}{person} & \rotatebox{60}{plant} & \rotatebox{60}{skbrd} & \rotatebox{60}{train} & \rotatebox{60}{truck}  & \rotatebox{60}{unk} & \rotatebox{60}{Knwn} & \rotatebox{60}{Mean}  &  \rotatebox{60}{Overall}\\
 \Xhline{0.8pt} 



Source& syn & real&36&27&21&49&66&0&69&1&42&8&59&0&81&31&	35& 57  \\
Source*& syn & real& 49&20&29&47&62&27&79&3&37&19&70&1&62&36&	38&50 \\
MMD& syn & real&48 & 49 & 28 & 59 & 77 & 2 & 82 & 5 & 43 & 16 & 64 & 3 & 82 & 39 & 42&62\\
MMD*& syn & real&51&40&42&56&68&24&75&2&39&30&71&2&75&41&44&59\\
DANN& syn & real& 53&5&31&61&75&3&81&11&63&29&68&5&76&43&40&60\\
DANN*& syn & real&59&41&16&54&77&18&88&4&44&32&68&4&61&42&43 &52\\
SE & syn & real & 94 & 82 & 87 & 67 & 94 & 31 & 91 & 63 & 89 & 76 & 80 & 33 & 53&73&72& 64\\
AODA&syn&real&85&71&65&53&83&10&79&36&73&56&79&32&87&60&62&73\\


\hline
Oracle &real& real& 96 & 81 & 81 & 82 & 92 & 69 & 92 & 64 & 86 & 84 & 92 & 67 & 91 & 82 & 82& 86\\
\Xhline{0.8pt}
\end{tabular}

\textit{\textbf{Known-to-Unknown Ratio = 1:10}}

\begin{tabular}{p{1cm}| p{0.4cm}p{0.5cm}|p{0.45cm} p{0.45cm} p{0.45cm} p{0.45cm} p{0.45cm} p{0.45cm} p{0.45cm} p{0.45cm} p{0.45cm} p{0.45cm} p{0.45cm} p{0.45cm} p{0.45cm} !{\vrule width0.8pt} p{0.6cm} p{0.6cm} p{0.6cm} }
\Xhline{0.8pt}
Source&syn&real&23&24&43&40&44&0&56&2&24&8&47&1&93&26&31&86\\
SE&syn & real& 94 & 74 & 86 & 68 & 91& 26 & 95&46 & 85 & 40 & 79 & 11 & 51 & 66 &65 &53\\
AODA &syn&real&80&63&59&63&83&12&89&5&61&14&79&0&69&51&52&68\\
\hline
Oracle&real&real&97&91&93&89&95&34&95&65&91&86&94&63&100&83&84&98\\
\Xhline{0.8pt}
\end{tabular}

} 

\vspace{-0.2cm}
\caption{\scriptsize Accuracy (\%) of experiments on openset classification dataset. The asterisk (*) indicates that we used Open set SVM for a classifier and therefore did not use existing ``known unknown'' source samples. \textit{Knwn} indicates the mean of class accuracies for known classes whereas \textit{Mean} also includes unknown class. \textit{Overall} indicates the mean accuracy computed across all \textit{samples} in the dataset. The ``unk'' category dominates the openset dataset and therefore low accuracy for unk has high impact on Overall accuracy. The table below shows that when the Known-to-Unknown is set to 1:10, the experimental results drop significantly for SE and AODA model. 
\vspace{-0.8cm}}
\label{tab_cls_opn}
\end{table}

%% file: 6_conclusion.tex
\section{Conclusion}

In this paper, we introduce three large scale synthetic-to-real datasets for different tasks within unsupervised domain adaptation all sharing same domain shift: object classification, open-set classification and object detection. We also present analysis of current state of the art in synthetic-to-real adaptation and explore how different properties of this setup, such as size of images, amount of training data, pre-training and background clutter influence adaptation performance.



We opensourced all resources and tools discussed in this paper to enable the community to develop and test novel domain adaptation models using an established protocol. Visit http://ai.bu.edu/syn2real for more details.



\section{Acknowledgements}
This work was supported in part by the Intelligence Advanced Research Projects Activity (IARPA) via Department of Interior/ Interior Business Center (DOI/IBC) contract number D17PC00341\footnote{\scriptsize The U.S. Government is authorized to reproduce and distribute reprints for Governmental purposes notwithstanding any copyright annotation thereon. Disclaimer: The views and conclusions contained herein are those of the authors and should not be interpreted as necessarily representing the official policies or endorsements,   either expressed or implied, of IARPA, DOI/IBC, or the U.S. Government}, by the National Science Foundation (NSF) via awards IIS-1212928, CNS-1629700, and CCF-1723379, and by the Honda Research Institute.

%% file: 7_supplementary.tex
\section{Supplementary Appendix}
\renewcommand{\thesubsection}{\Alph{subsection}}

\subsection{Other relevant datasets}
In Table~\ref{tab_datasets} we list the existing cross-domain datasets used for open- and closed-set object classification and object detection. We display other relevant cross-domain datasets in Table~\ref{tab_datasets_extra}. Table~\ref{tab_synth_datasets} shows the comparison between \Name-C and other synthetic object datasets.

\begin{table}[h]
    \centering
    DIGIT CLASSIFICATION\\
    \begin{tabular}{|c|c c c|}\hline
    Dataset & Examples & Classes & Domains \\ \hline
    USPS-subset \cite{hull1994database} & 1,800     & 10& 1 \\
    MNIST-subset \cite{lecun1998gradient} & 2,000     & 10& 1 \\
    USPS-Full \cite{hull1994database}       & 9,000     & 10& 1 \\
    MNIST-Full \cite{lecun1998gradient}  & 70,000    & 10& 1 \\
    SVHN \cite{netzer2011reading} & 630,420   & 10& 1 \\ \hline
    \end{tabular}
~\\ 
    \vspace{.2cm}
    FACE RECOGNITION \\
    \begin{tabular}{|c|c c c|} \hline
    Dataset & Examples & Classes & Domains \\ \hline
    PIE \cite{sim2002cmu}  & 11,554    & 68& 1 \\ \hline 
    \end{tabular} \\
~\\ 
    \vspace{.1cm}
    SEMANTIC SEGMENTATION \\
    \begin{tabular}{|c|ccc|} \hline
    Dataset & Examples & Classes & \ Domains \ \\ \hline
    SYNTHIA-subset\cite{RosCVPR16} &  9,400 & 12 & 1  (city)\\ 
    CityScapes \cite{Cordts2016Cityscapes} & 5,000 & 34 & 1  (city)\\
    GTA5 \cite{richter2016playing} & 24,966 & 18 & 1  (city)\\ \hline
    \end{tabular} \\
    \vspace{.3cm}
    \caption{Other popular domain adaptation datasets that are not directly compatible with \Name ~in domain or task.} 
    \label{tab_datasets_extra}
\end{table}    

\begin{table}[h]
    \centering
    SYNTHETIC OBJECTS \\
    \begin{tabular}{|c|c c c|} \hline
    Dataset & Models & Images & Classes \\ \hline
    ModelNet \cite{wu20153d} & 127,915 & - & 662 \\
    PASCAL3D+ \cite{xiang2014beyond} & 77 & 30,899 & 12 \\
    ObjectNet3D \cite{xiang2016objectnet3d}& 44,147 & - & 100 \\
    ShapeNet-Core\cite{shapenet2015}& 51,300 & - & 55 \\
    ShapeNet-Sem\cite{shapenet2015}& 12,000 & - & 270 \\   Redwood\cite{Choi2016}& 10,000 & - & 44 \\
    IKEA \cite{lpt2013ikea} & 219 & - & 11  \\ \hline
    \textbf{\Name-C (source)}   & 1,907 &  152,397 & 12 \\ \hline
    \end{tabular} \\
    \vspace{.3cm}
    \caption{Comparison between \Name-C and existing synthetic object datasets. The majority of 3D model databases have thousands of rare classes with significantly unequal number of samples that are not present in the majority of other datasets that makes it difficult to use them for cross-domain adaptation. At the same time, \Name-C dataset delivers a substantial balanced collection of models with all metadata (orientation, scaling, etc.) required for proper rendering and a deliberately limited number of classes mostly overlapping with ``standard" VOC PASCAL classes.}
    \label{tab_synth_datasets}
\end{table}

\begin{figure}
    \centering
    \includegraphics[width=0.3\linewidth]{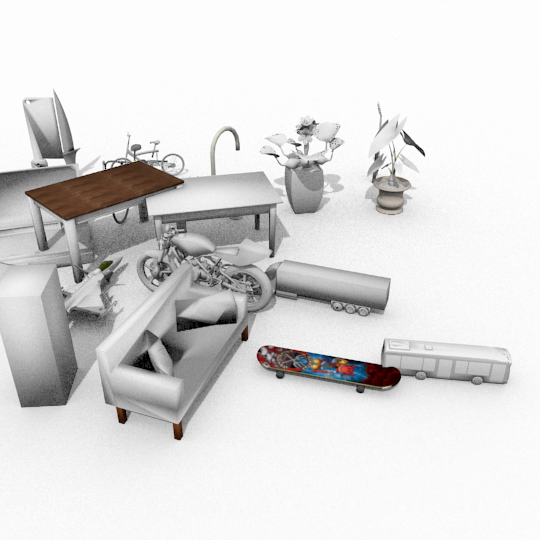}
    \includegraphics[width=0.3\linewidth]{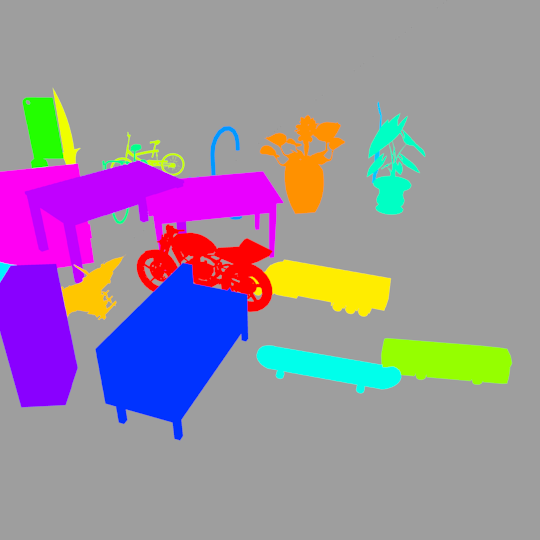}
    \includegraphics[width=0.3\linewidth]{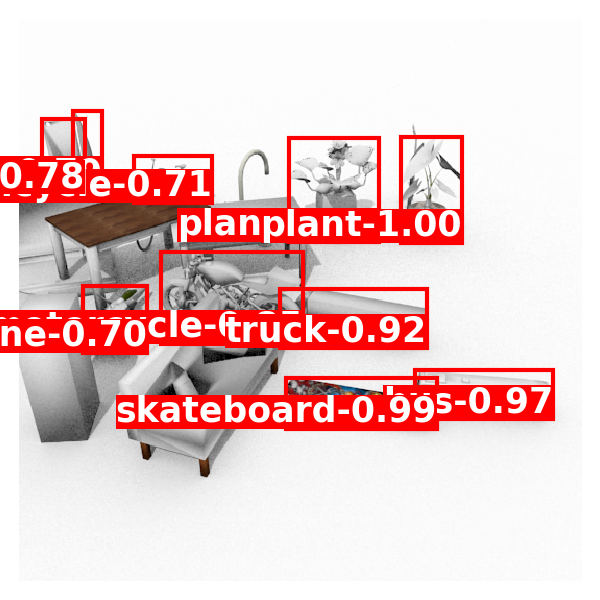}
    \includegraphics[width=0.3\linewidth]{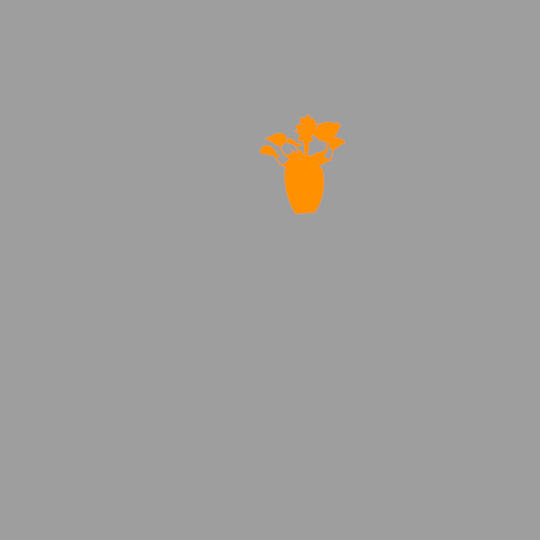}
    \includegraphics[width=0.3\linewidth]{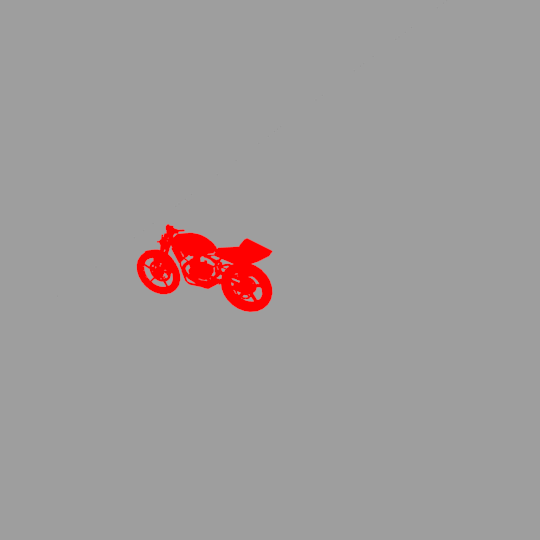}
    \includegraphics[width=0.3\linewidth]{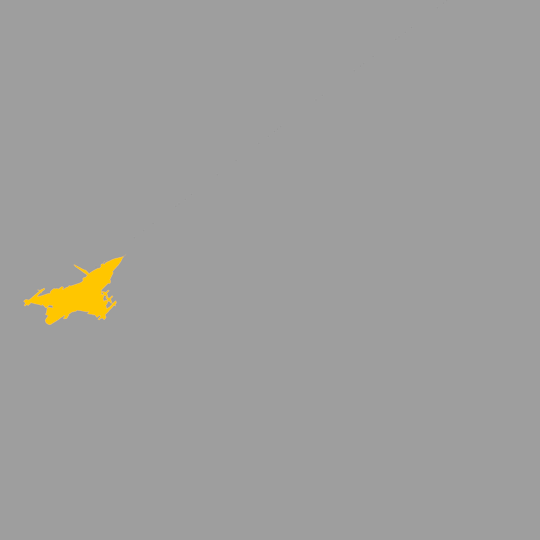}
    \caption{We generated bounding boxes and occlusion rates by comparing instance segmentation maps with instance segmentation of the same scene with only single object present. We obtain segmentation maps by replacing material shader with a constant value shader.}

    \label{fig:my_label}
\end{figure}

\subsection{\Name~  Image samples}

\begin{figure*}
\centering
\includegraphics[width= 1\linewidth]{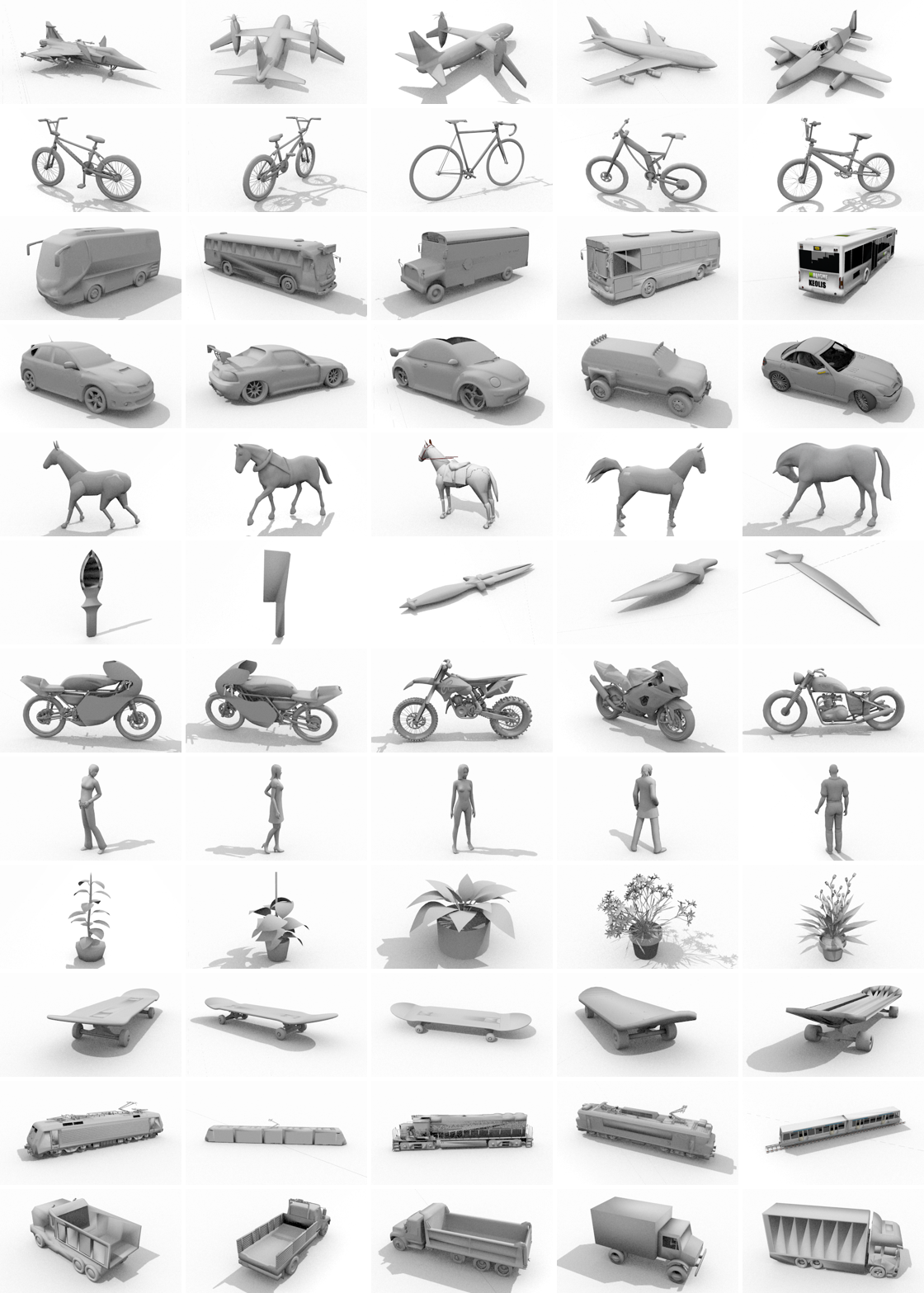}
\caption{\textbf{Training domain sample data from \Name-C.} Images generated from Synthetic CAD models.}
\label{class_train_data}
\end{figure*}

\begin{figure*}
\centering
\includegraphics[width= 1\linewidth]{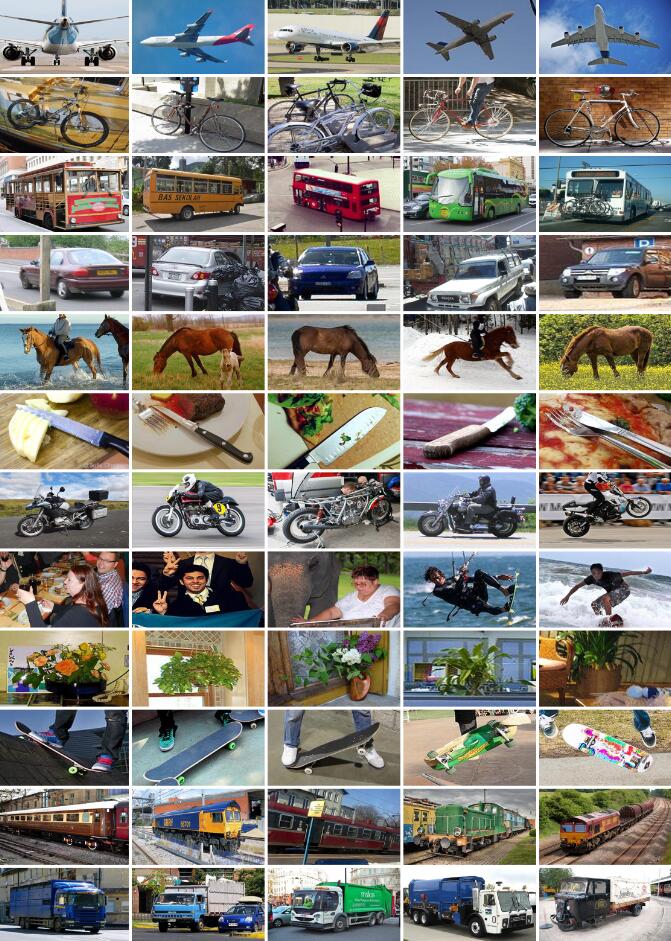}
\caption{\textbf{Validation domain sample data from \Name-C.} Real images cropped from the MS COCO dataset.}
\label{class_val_data}
\end{figure*}

\begin{figure*}
\centering
\includegraphics[width= 1 \linewidth]{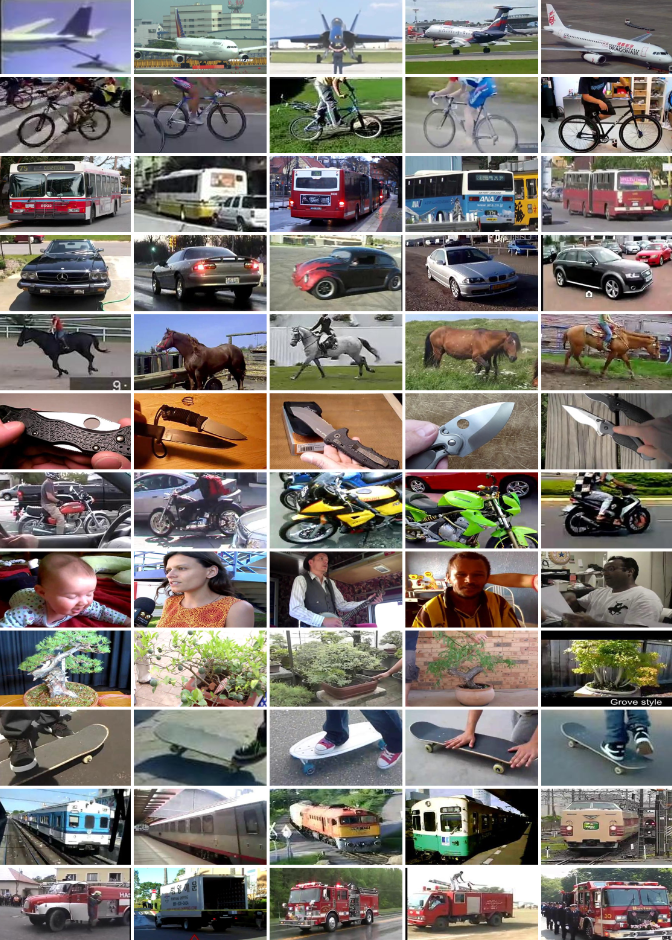}
\caption{\textbf{Test domain sample data from \Name-C.} Real images from YouTube-BB dataset.}
\label{class_test_data}
\end{figure*}

\begin{figure*}
\centering
\includegraphics[width= 1 \linewidth]{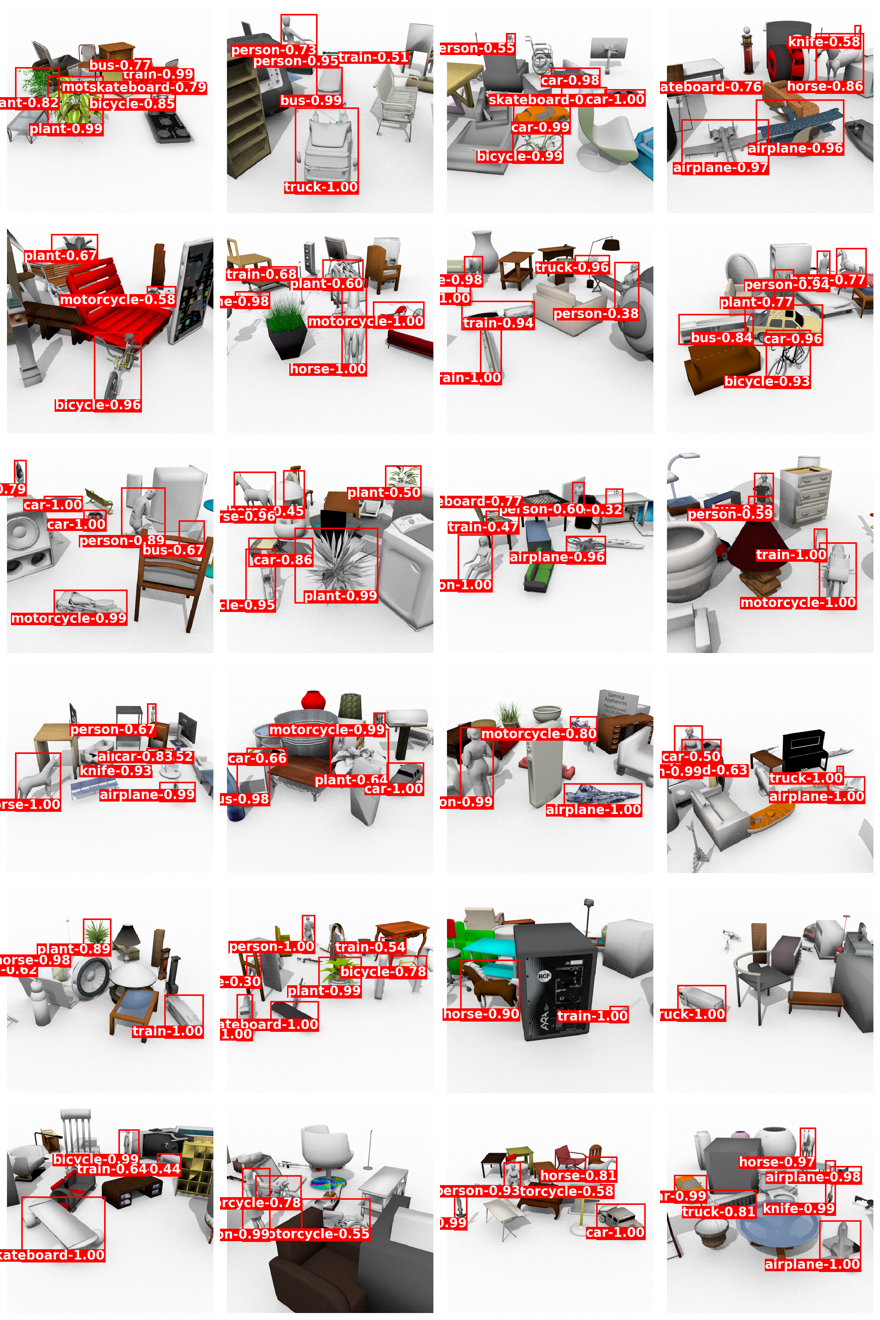}
\caption{\textbf{Training domain sample data from \Name-D.} Synthetic images generated from CAD models. Numbers indicate occlusion rates for specific objects.}
\label{detection_train_data}
\end{figure*}

\begin{figure*}
\centering
\includegraphics[width= 1 \linewidth]{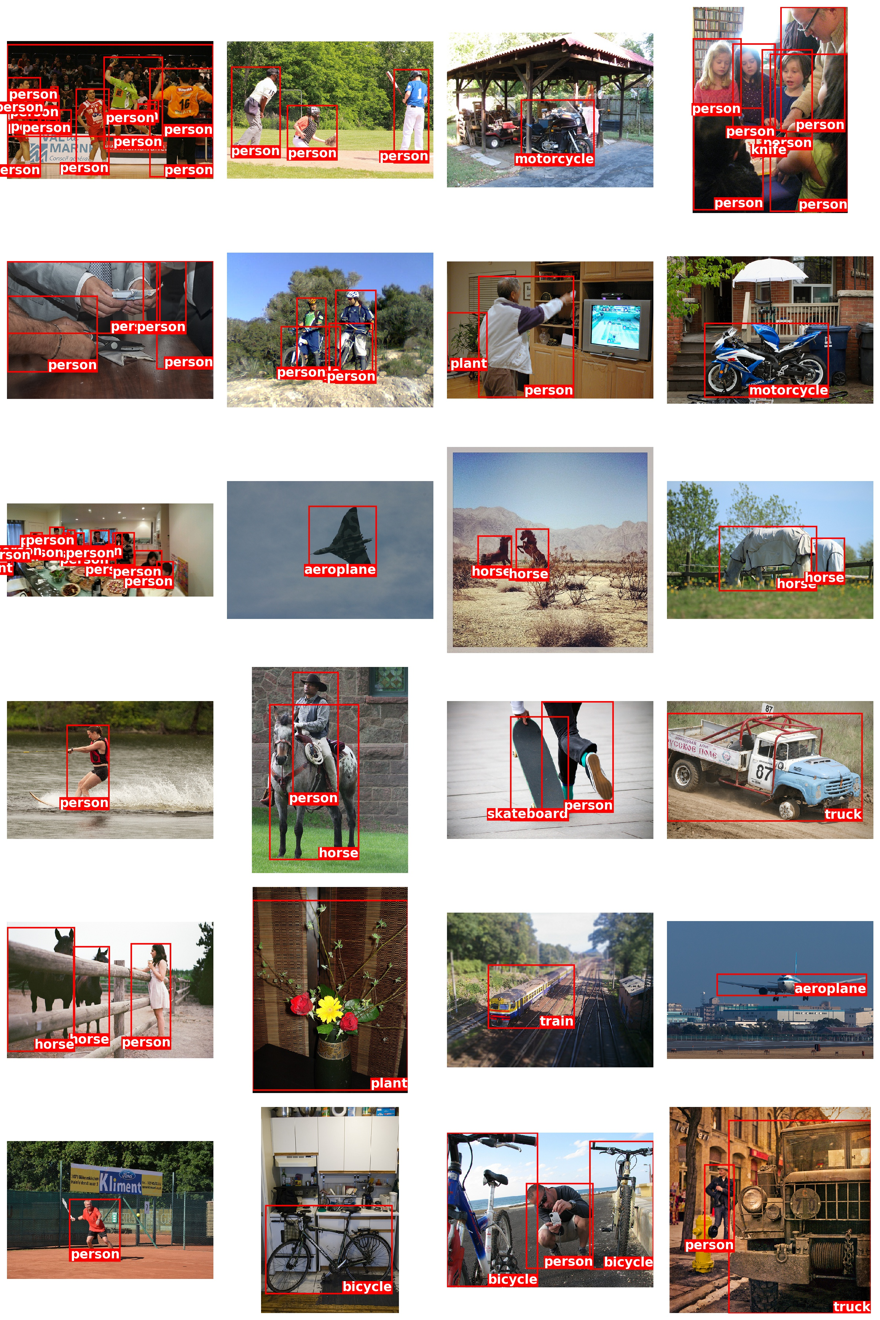}
\caption{\textbf{Validation domain sample data from \Name-D.} Real images from the MS COCO dataset.}
\label{detection_val_data}
\end{figure*}

\begin{figure*}
\centering
\includegraphics[width= 0.95 \linewidth]{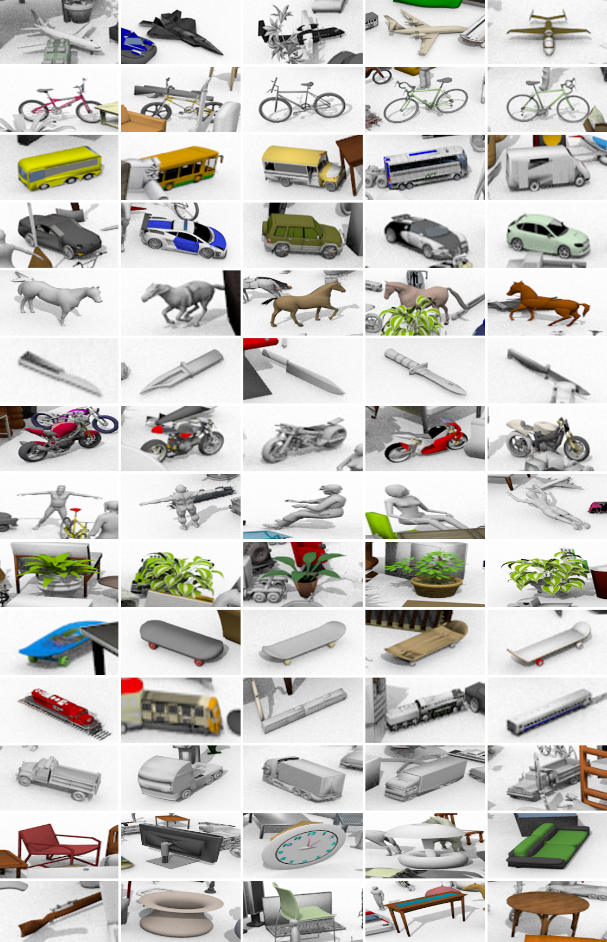}
\caption{\textbf{Training domain sample data from \Name-O.} The first 12 rows show the images of ``known'' category and the last two rows show the images of ``unknown'' category.}
\label{openset_train_data}
\end{figure*}

\begin{figure*}
\centering
\includegraphics[width= 1 \linewidth]{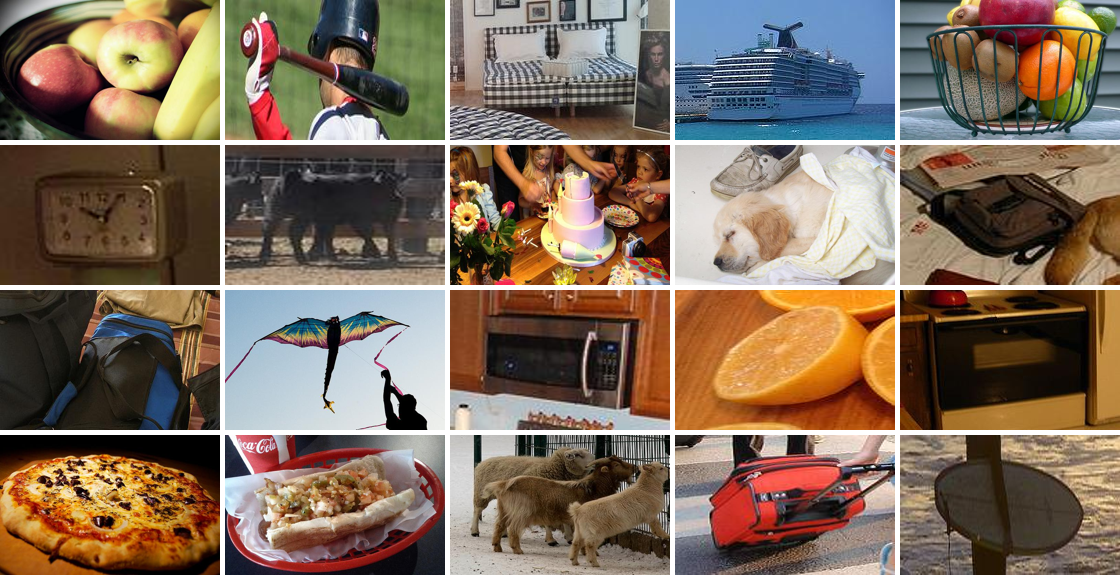}
\caption{\textbf{``Unknown'' object sample data in validation domain from \Name-O.} The images of ``known'' category are the same as images in close set \Name-C \ie Figure \ref{class_val_data}.  }
\label{openset_val_data}
\end{figure*}

We presented images from the \Name~ dataset in Figures~\ref{fig_cls_sample}. More images of each considered category for all known domains are given in Figures~\ref{class_train_data}--\ref{openset_val_data}. 


\clearpage

\clearpage

\clearpage

\clearpage

\clearpage

\clearpage

\clearpage
